# Deep Embedding Convolutional Neural Network for Synthesizing CT Image from T1-Weighted MR Image

Lei Xiang[1], Qian Wang[1,*], Xiyao Jin[1], Dong Nie[3], Yu Qiao[2], Dinggang Shen[3,4,*]

[1]Med-X Research Institute, School of Biomedical Engineering, Shanghai Jiao Tong University, Shanghai, China

[2]Shenzhen Key Lab of Comp. Vis. & Pat. Rec., Shenzhen Institutes of Advanced Technology, CAS, Shenzhen, China

[3]Department of Radiology and BRIC, University of North Carolina at Chapel Hill, NC, USA

[4]Department of Brain and Cognitive Engineering, Korea University, Seoul 02841, Republic of Korea

*Corresponding authors: wang.qian@sjtu.edu.cn, dgshen@med.unc.edu*

**Abstract**

Recently, more and more attention is drawn to the field of medical image synthesis across modalities. Among them, the synthesis of computed tomography (CT) image from T1-weighted magnetic resonance (MR) image is of great importance, although the mapping between them is highly complex due to large gaps of appearances of the two modalities. In this work, we aim to tackle this MR-to-CT synthesis by a novel deep embedding convolutional neural network (DECNN). Specifically, we generate the feature maps from MR images, and then transform these feature maps forward through convolutional layers in the network. We can further compute a tentative CT synthesis from the *midway* of the flow of feature maps, and then embed this tentative CT synthesis back to the feature maps. This embedding operation results in better feature maps, which are further transformed forward in DECNN. After repeating this embedding procedure for several times in the network, we can eventually synthesize a final CT image in the end of the DECNN. We have validated our proposed method on both brain and prostate datasets, by also comparing with the state-of-the-art methods. Experimental results suggest that our DECNN (with repeated embedding operations) demonstrates its superior performances, in terms of both the perceptive quality of the synthesized CT image and the run-time cost for synthesizing a CT image.

## 1   Introduction

Computed tomography (CT) and structural magnetic resonance (MR) images are both important and widely applied in various medical cases, e.g., detecting infarction and tumors in head (Sze et al., 1990; Williams et al., 2001), detecting both acute and chronic changes in lung parenchyma (Remy-Jardin et al., 1996; Schwickert et al., 1994), and imaging complex fractures in extremities (Haidukewych, 2002; Soto et al., 2001). Recently, it has become desirable that a CT image can be synthesized from the corresponding MR scan. For example, quantitative positron emission tomography (PET) requires attenuation correction (Carney et al., 2006; Kinahan et al., 1998; Pan et al., 2005). Unlike the traditional PET/CT scanner, the MR signal in cutting-edge PET/MR is not directly correlated to tissue density and thus cannot be applied to attenuation correction after a simple intensity transform (Wagenknecht et al., 2013). As a possible solution, one may have to conduct image segmentation, which is challenging to implement though, to reveal tissue distribution first. In this case, the challenge can be greatly alleviated with the CT image synthesized from the T1-weighted MR image.

In this work, we aim to address the problem of synthesizing CT from T1 MR. But these two modalities differ largely regarding their image appearances, which makes this synthesis problem challenging to solve. Examples of the T1-weighted MR images and the corresponding CT images are shown in Fig. 1. These images are acquired from the same patients, i.e., (a) for the brain, and (b) for the prostate, respectively. In the MR images, the intensity values of 'air' and 'bone', as pointed by the blue arrows and orange arrows, are both low. However, in the CT images, the 'air' appears dark, while the 'bone' turns to be bright. In general, the intensity mapping between these two modalities of MR and CT is highly complex, encoding both spatial and context information in non-linear mapping.

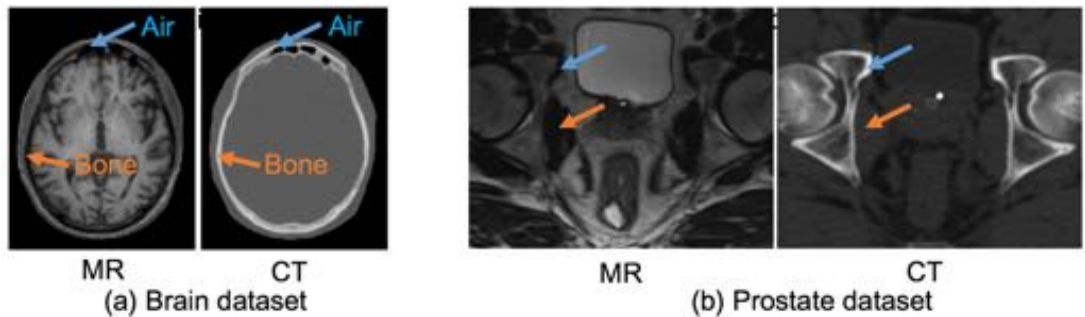

**Fig. 1.** Examples of the pairs of MR and CT data from the same patients: (a) brain, and (b) prostate. The inter-modality mapping is highly complex and non-linear, due to inconsistent appearance mapping in the two modalities, i.e., dark air in both MR and CT, while dark bone in MR but bright bone in CT.

There are several reports in the literature focusing on inter-modality medical image synthesis, i.e., from MR to CT. These methods can be mainly categorized into the following three classes.

(a) **Atlas-based methods**. In the atlas-based methods (Arabi et al., 2016; Hofmann et al., 2008; Kops and Herzog, 2007), a set of atlases are prepared in advance, each of which consists of both MR and CT acquisitions. Given a new subject with the MR image only, all atlases are first registered with the new subject by referring to their respective MR images. Then, the resulted deformation fields are used to warp the respective CT images of the atlases to the new subject space, from which the subject CT image can be synthesized through fusion of aligned atlas CT images (Burgos et al., 2014). Clearly, the performance of the above methods is highly related with the registration accuracy, and the synthesized CT image for the new subject could possibly suffer from large variation to the atlas and inaccurate estimation of the deformation. Note that the atlas-based method may also cost high in time for registering all images.

(b) **Sparse-coding-based methods**. These methods (Yang et al., 2012; Yang et al., 2008) usually involve several steps in the respective pipelines. First, the overlapping patches are extracted from the new subject MR image. These subject MR patches are then encoded by a MR patch dictionary that is built from the linearly aligned MR atlases. The obtained sparse representation coefficients are transferred to the coupled CT patch dictionary (also built from the linearly aligned CT atlases), to fuse the respective CT atlas patches for finally synthesizing the subject CT image. (Roy et al., 2010) applied this framework for predicting FLAIR image from $T_1$- and $T_2$-weighted MR images. Similarly, (Ye et al., 2013) estimates $T_2$- and diffusion-weighted MR images from $T_1$-weighted MR. But one main drawback of these methods is that the estimation is computationally expensive (Dong et al., 2016a) due to the need of sparse coding optimization upon all image locations. Because each location needs extract patch and go through all the operations to get the corresponding predicted patch. And constructing a global dictionary means the dictionary is of big size to ensure the final prediction performance, which obviously add the cost time for solving the sparse representation coefficients (Dong et al., 2016a; Yang et al., 2008).

(c) **Learning-based methods**. These methods learn the complex mapping from the local detailed appearances of MR images to those of CT images in the same subjects (Huynh et al., 2016; Johansson et al., 2011; Roy et al., 2014). In order to address the issue of expensive computation in sparsity learning based methods, (Huynh et al., 2016) presented an approach to estimate CT image from MR using the structured random forest and the auto-context model. (Vemulapalli et al., 2015) proposed an unsupervised approach to maximize both global mutual information and local spatial consistency for inter-modality image synthesis. But, such methods often have to first decompose the whole input MR image into the overlapping patches, and then extract features for each patch. After mapping the MR-related features to the (output) CT image patch, one has to take an additional step to integrate all estimated CT patches for synthesizing the entire CT image. Therefore, these methods suffer from the complexity from two aspects: multiple steps to get the mapping from MR to CT and post-processing by patch-by-patch integrating.

Recently, the convolutional neural network (CNN) has shown its tremendous popularity and good performance in the computer vision and medical image computing fields. CNN is capable of modeling non-linear mapping between different image spaces, without defining the hand-crafted features. And CNN-based method can overcome time-consuming problem of the patch-based method by taking whole image as input and output its whole image prediction during testing stage. Successful applications can be found by reconstructing high-resolution images from low-resolution images (Dong et al., 2016a), and by enhancing PET signals from the simultaneously acquired structural MR (Li et al., 2014). (Han, 2017) also proposed a deep convolutional neural network method for CT synthesis from MR image, which achieved reasonable performance compared to the atlas-based methods. However, this method can only process a single slice through each forward mapping. To handle the problem of 3D MR-to-CT synthesis, this method had to process multiple slices independently, which could often cause discontinuity and artifacts in the synthesized CT images.

In this paper, we propose a deep *embedding* convolutional neural network (DECNN) to synthesize CT images from T1-weighted MR images. Concerning the examples in Fig. 1, the mapping from MR to CT can be highly complex, as the appearances of these two modalities vary significantly across spatial locations (Wagenknecht et al., 2013). This large inter-modality appearance gap challenges the accurate learning of CNN. To this end, we decompose the CNN model into two stages: the *transform stage* and the *reconstruction stage*. The *transform stage* is a collection of the convolutional layers that are responsible for forwarding the feature maps, while the *reconstruction stage* aims to synthesize the CT image from the transformed feature maps. Besides, we also propose a novel *embedding block*. The

*embedding block* is able to synthesize the CT image from the tentative feature maps in the CNN. Next, the tentative CT synthesis is embedded with the feature maps, thus the newly embedded feature maps become more related to the CT images and can be further refined by the subsequent layers in the CNN. More importantly, we insert multiple *embedding blocks* into the *transform stage* to derive our DECNN accordingly. The embedding block is similar to deep supervision (Lee et al., 2015) which has been adopted by many computer vision tasks (Chen et al., 2016; Xie and Tu, 2015). Holistically-nested edge detection (HED) method (Xie and Tu, 2015), for example, leverages multi-scale and multi-level feature learning to perform image-to-image edge detection. This method results in multi-outputs and fuses them in the end of the network. DCAN (Chen et al., 2016) takes advantages from the auxiliary supervision by introducing multi-task regularization during training. Our embedding block goes beyond deep supervision as the midway feature maps are further embedded into the subsequent layers of the network. The embedding block thus provides consistent supervision to facilitate the modality synthesis and improve the quality of the final results. This embedding strategy also resembles the auto-context (Tu and Bai, 2010) strategy. We note that auto-context generally requires independent learning for each stage, while our DECNN can integrate all embedding blocks into a unified network for end-to-end training.

It is worth indicating that the introduction of the *embedding block* has greatly strengthened the inter-modality mapping capability of DECNN regarding MR and CT images. In particular, the tentatively synthesized CT images are embedded into better feature maps, which will be transformed forward to refine the synthesis of CT image. Also, this *embedding block* contributes to the convergence of the training of this very deep network in a back-propagation form. Moreover, DECNN allows us to process all test subjects in a very efficient end-to-end way.

Our main contributions can be summarized as follows.
- We propose a very deep network architecture for estimating CT images from MR images directly. The network consists of convolutional and concatenation operations only. It can thus learn an end-to-end mapping between different imaging modalities, without any patch-level pre- or post-processing.
- To better train the deep network and refine the CT synthesis, we propose a novel embedding strategy, to embed the tentatively synthesized CT image into the feature maps and further transform these features maps forward for better estimating the final CT image. This embedding strategy helps back-propagate the gradients in the network, and also make the training of the end-to-end mapping from MR to CT much easier and more effective.
- We carry out experiments on two real datasets, i.e., human brain and prostate datasets. The experimental results show that our method can be *flexibly adapted* to different applications. Moreover, our method outperforms the state-of-the-art methods, regarding both the accuracy of estimated CT images and the speed of synthesis process.

The rest of this paper is organized as follows. In Section 2, we present the details of our proposed DECNN for estimating CT image from MR image. Then, in in Section 3, we conduct extensive experiments, evaluated with different metrics, on both real brain and prostate datasets. Finally, we conclude this paper in Section 4.

## 2   Method

CNN is capable of learning the mapping between different image spaces. We adopt the CNN model similar to (Dong et al., 2016a) for the task of MR-to-CT image synthesis, and then develop our DECNN accordingly. As mentioned above, we decompose the CNN model into two stages, i.e., the *transform stage* and the *reconstruction stage*, as also illustrated in Fig. 2 (a). The *transform stage* is used to forward the feature maps (i.e., derived from MR images), such that the CT image can be synthesized in the *reconstruction stage*. Our proposed DECNN is featured by the *embedding block*, which embeds the tentative synthesis of the CT image into the feature maps. This *embedding block* is a key factor of our DECNN for more accurately mapping from T1-weighted MR images to CT images.

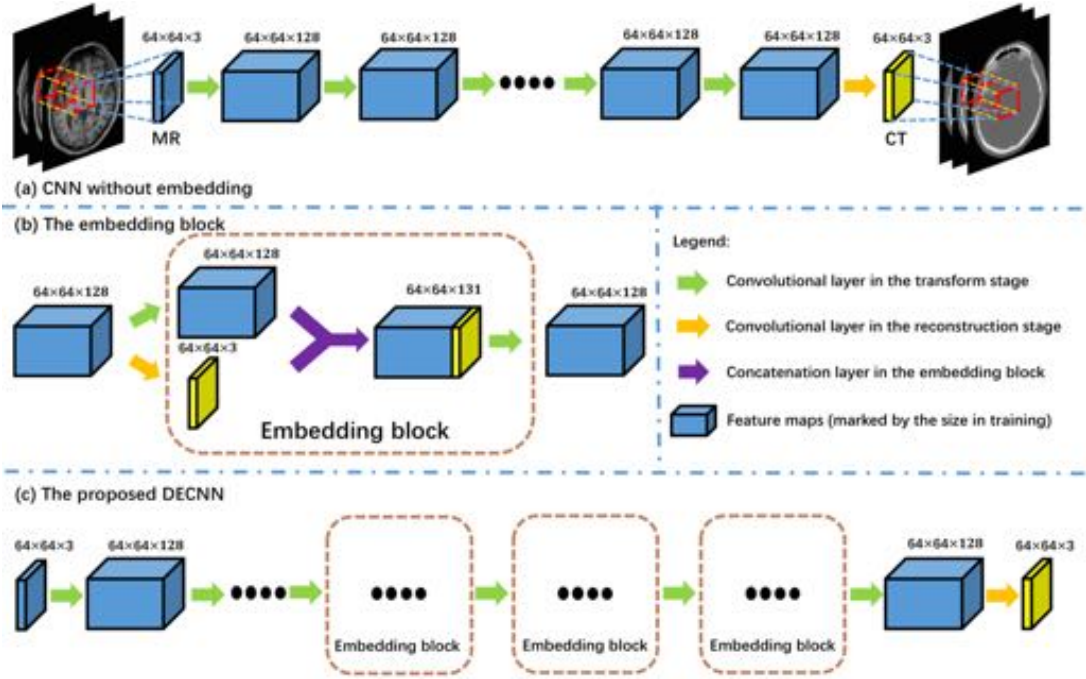

**Fig. 2.** Illustrations of (a) the CNN model *without* embedding, (b) the embedding block, and (c) our proposed DECNN. The CNN model in (a) consists of the transform stage and the reconstruction stage. Our proposed DECNN model shown in (c) is derived by inserting multiple embedding blocks into the CNN model in (a). The sizes of the feature maps are marked for the networks during the training. Note that the marked axial size (64×64) is not required when testing on a new subject.

### 2.1   Transform Stage and Reconstruction Stage

**Transform Stage:** The transform stage in CNN consists of several convolutional layers, which derive the feature maps from the input MR images and forward them for later CT synthesis in the reconstruction stage. Denoting the input feature maps to the $j$-th convolutional layer as $F_{j-1}$ (corresponding to the output of the early layer, the $(j-1)$-th layer), we can then compute the output of the $j$-th layer as

$$F_j = \sigma(W_j * F_{j-1} + B_j), \tag{1}$$

where $W_j$ and $B_j$ represent the kernels associated with the $j$-th convolutional layer, and '*' denotes the convolution operator. We choose the parametric rectified linear unit (PReLU) (He et al., 2015b) as the activation function due to its outstanding generalization performance in similar tasks (Dong et al., 2016b; Hui et al., 2016). Here, $\sigma$ is defined as:

$$\sigma(\mathbf{x}) = \max(0, \mathbf{x}) + \boldsymbol{\alpha} \min(0, \mathbf{x}), \tag{2}$$

where $\boldsymbol{\alpha}$ is a learnable parameter.

**Reconstruction Stage:** The CNN model ends with the reconstruction stage that synthesizes CT images from the feature maps outputted by the last layer of the transform stage. The reconstruction can be attained by a single convolutional layer by following

$$\boldsymbol{F}_R = \boldsymbol{W}_R * \boldsymbol{F}_T + \boldsymbol{B}_R. \tag{3}$$

Here, $\boldsymbol{F}_T$ is the feature map outputted by the transform stage, and $\boldsymbol{F}_R$ is the synthesis estimated by the reconstruction stage. Note that there is no activation function employed in the reconstruction stage. We also require the size of $\boldsymbol{F}_R$ to be equal with the size of the initial input to the transform stage.

The training of the convolutional layers is conducted in a patch-by-patch way in order to alleviate the demand of many subjects for training. Given a center location in a training subject with his/her carefully aligned MR and CT images, we extract the corresponding 64×64×3 surrounding MR and CT patches, from 3 consecutive axial slices. The MR patch can thus be perceived as 3 feature maps, each with size of 64×64, and are input to the first convolutional layer of the network. Specifically, 128 feature maps are generated from these initial input MR patch and forwarded across all convolutional layers in the transform stage. To this end, there are 128 kernels in each convolutional layer, while the support of each kernel is 3×3×128 (and 3×3×3 for the first convolutional layer only). Our experiments show that the 3×3 *axial* receptive field of the kernel is enough to complete the mapping from MR to CT. Note, we apply zero-padding such that the sizes of all feature maps are always 64×64 in the transform stage. In the final, the transform stage yields 128 feature maps of the size 64×64, from which the CT patch of the size 64×64×3 can be synthesized in the reconstruction stage.

Note that a single convolutional layer is needed in the reconstruction stage. We particularly adopt 3 kernels of the support 3×3×128 to complete the reconstruction. The output of each kernel is associated with a certain 64×64 axial slice in the synthesized CT patch. Note that the 64×64 axial size is only required upon the training patches. In the testing phase, the network can directly apply convolutional operations upon 3 consecutive axial MR planes of arbitrary size. Thus, the entire CT field-of-view can be synthesized efficiently by the end-to-end mapping of CNN.

## 2.2 Embedding Block and DECNN

The *embedding block* estimates the CT synthesis given the latest feature maps, and then embeds the tentatively synthesized CT image into better feature maps. The architecture of the *embedding block* is shown in Fig. 2 (b), as stated before. The *embedding block* consists of 4 steps.

(1) The CT image corresponding to the input MR image is tentatively synthesized from the feature maps that are input to the embedding block under consideration. The synthesis is accomplished by applying Eq. (3) to the input feature maps; this step is the same as the reconstruction stage.

(2) The input feature maps are transformed through a single convolutional layer in the *embedding block*. The number of the kernels and the support of each kernel are the same as the convolutional layers in the transform stage, which has been described in Section 2.1.

(3) The tentatively synthesized CT image is perceived as 3 additional feature maps, and then concatenated with those 128 feature maps yielded by Step 2 to obtain 131 feature maps.

(4) These 131 concatenated feature maps are transformed through a subsequent convolutional layer. This layer reduces the number of the feature maps from 131 to 128, implying that there are 128 kernels of the support $3\times 3\times 131$ in this layer.

In general, the feature maps are processed through two convolutional layers and one concatenation layer in the *embedding block*. The sizes of the input and the output feature maps for the *embedding block* are identical. Thus, the *embedding block* can be seamlessly integrated into the aforementioned CNN model.

We insert several *embedding blocks* in the late part of the *transform stage*, and thus derive our proposed DECNN model as shown in Fig. 2 (c). Note that, after each embedding block, an additional convolutional layer is appended to refine the feature maps. Finally, the feature maps arrive at the *reconstruction stage*, where the CT image is eventually synthesized. In general, with $k$ embedding blocks, the total number of the convolutional layers in DECNN will be $3k + 5$.

Note that the *embedding blocks* are essentially important to tackle the highly complex mapping from MR to CT. The deep network with many convolutional layers transforms feature maps forward. The representations captured by *shallow layers* are more related with the input MR appearances, while the representations captured by *deep layers* are more related with the CT appearances. The *embedding blocks* in the late part of the *transform stage* can first synthesize CT images tentatively, and then inject the tentative CT estimation into the feature maps for refined CT synthesis. In this way, these *embedding blocks* help fill the large appearance gap between the two modalities and also speeds up the MR-to-CT mapping within the flow of feature maps. The CT synthesis can also be refined across individual *embedding blocks*. Therefore, the learning of DECNN tends to yield more accurate and effective modeling for the mapping from MR to CT.

## 2.3 Implementation Details

The introduction of the *embedding blocks* also allows us to compute the auxiliary loss, which helps improve the robustness of DECNN against the vanishing gradients and also strengthens the learning of the network parameters during back-propagation. For the $k$-th embedding block in DECNN, the tentatively synthesized CT image is denoted by $F_{E_k}$. Concerning the ground truth $\widetilde{F}$, we can compute the overall loss function of DECNN as

$$L(\boldsymbol{\theta}) = \|F_R - \widetilde{F}\|^2 + \beta \sum_k \|F_{E_k} - \widetilde{F}\|^2 + \alpha\, \varphi(\boldsymbol{\theta}). \tag{4}$$

Here, $\boldsymbol{\theta}$ encodes all parameters of the network, and $\varphi(\boldsymbol{\theta})$ is the $L_2$-norm regularization upon the parameters. We set $\beta = 0.5$ in our experiments, which balances the losses computed from the tentative CT estimation of individual embedding blocks and also the output of the final reconstruction stage, and $\alpha = 0.001$ for the regularization term.

In DECNN, we transform the input feature maps through 5 convolutional layers empirically, such that the feature maps are ready for the first embedding block to synthesize CT images tentatively. More embedding blocks can be flexibly appended, according to the need in different applications, as described in Section 3 below.

Our network processes the input MR patches covering 3 axial slices, and estimates the output CT patches of the same size. In the training, the axial size is fixed to 64×64 for convenient implementation. In the testing, the 3 consecutive axial slices, covering the entire slices, can be processed as a whole. Therefore, each axial slice of the subject CT image can be reconstructed 3 times, which will be averaged for the final CT synthesis. Note that more axial slices can be considered simultaneously, although the learning of DECNN may become more challenging. Note, with the 3 axial slices consecutively synthesized, the superior-inferior continuity of the synthesized CT image can be preserved, which will be confirmed by experimental results in Section 3.5.

## 3 Experimental Result

In this section, we evaluate the performance of our method on two real CT datasets, i.e., 1) brain dataset and 2) prostate dataset, which are the same dataset used in (Huynh et al., 2016; Nie et al., 2016). We first describe the datasets used for training and testing our method. Next, more detailed training setup is given. Subsequently, we analyze the effect of the *embedding blocks* in our architecture. We also present both qualitative and quantitative comparisons between our DECNN model and state-of-the-art methods. After that, we show the superiority of our designed quasi-3D mapping in keeping the appearance consistency in each dimension. Finally, we analyze the learned feature maps in our model, to explain why our DECNN model can iteratively get better synthesized CT results.

### 3.1 Datasets

We apply DECNN to the two real datasets for demonstrating its capability. The two datasets are introduced below.

(1) **Brain dataset**. The brain dataset consists of 16 subjects, each with both MR and CT scans, from the Alzheimer's Disease Neuroimaging Initiative (ADNI) dataset (www.adni-info.org). The CT images were acquired with a Siemens Somatom scanner. The voxel size of the images is $0.59\times0.59\times0.59mm^3$. The MR images were acquired using a Siemens Triotim scanner, with the voxel size $1.2\times1.2\times1mm^3$, TE 2.95 ms, TR 2300 ms, and flip angle $9°$.

(2) **Prostate dataset.** The prostate dataset consists of 22 subjects and . The voxel sizes of the CT and MR images are $1.17\times1.17\times1mm^3$ and $1\times1\times1mm^3$, respectively. The CT images were acquired using a Philips scanner, and the MR images were acquired by a Siemens Avanto scanner with TE 123 ms, TR 2000 ms, and flip angle $150°$.

In both datasets, CT images are aligned to the corresponding MR images through intra-subject registration and inter-subject registration, and CT images are resampled to the same size of MR images. The intra-subject registration means to align each pair CT and MR images of the same subject. For brain dataset, we apply FLIRT (Jenkinson and Smith, 2001) with 12 degrees of freedom to perform linear registration, because there is only small deformation in brain image and linear registration is sufficient. But for prostate dataset, there is often a large deformation on soft tissues. So we choose B-Splines (Klein et al., 2010) which uses mutual information as similarity measure to perform intra-subject deformable registration. Finally, rigid-body inter-subject registration is performed to roughly bring all subjects onto a common space. For the brain images, their backgrounds are removed and the final images have the size of 234×181×149. For prostate images, since only the pelvic region is concerned in this study, we crop the original images to be centered at the pelvic region, which leads to the size of 153×193×50 for each prostate image. Typical examples of the pre-processed MR and CT images from the brain dataset and the prostate dataset are shown in Fig. 1.

### 3.2 Experimental Setting

Various studies have found that, with more data, the performance of a deep model can get better (Dong et al., 2016a; Timofte et al., 2014). So, in order to make full use of the dataset, we adopt a data augmentation strategy similar to (Timofte et al., 2015; Wang et al., 2015). All images are left-right flipped. In this way, we are able to have more available images for training the deep model in each leave-one-out case. We implement the CNN-based learning with CAFFE (Jia et al., 2014) in this work. The kernels in the convolutional layers are initialized following (He et al., 2015a). Similar to (Hui et al., 2016; Jain and Seung, 2009), we set the learning rate to $10^{-5}$, and adopt "Adam" as the solver for the optimization during back-propagation. The momentum is set to 0.9, and the $L_2$-norm regularization upon the learned parameters is set to $10^{-3}$. The batch size is set to 16.

Instead of using large-size image for training, patches are generated by dividing each image into a regular grid of small overlapping patches. Specifically, we extract patches of the size 64×64 from the pre-processed MR and the ground-truth CT with the stride of 8. As we use leave-one-out strategy to verify our model, we can get enough samples for training such very deep architecture. Specifically, we extract almost 600000 patches for brain dataset, 100000 patches for prostate dataset.

To quantitatively evaluate the synthesis performance, we use two popular metrics: 1) mean absolute error (MAE) and 2) peak signal-noise ratio (PSNR). The MAE and PSNR are computed as

$$\text{MAE} = \frac{|\boldsymbol{F}_R - \widetilde{\boldsymbol{F}}|}{N}, \tag{5}$$

$$\text{PSNR} = 10 \ln \left( N \cdot \frac{Q^2}{\|\boldsymbol{F}_R - \widetilde{\boldsymbol{F}}\|^2} \right). \tag{6}$$

where $N$ is the number of voxels in each image, and $Q$ is the maximal intensity value of $\boldsymbol{F}_R$ and $\widetilde{\boldsymbol{F}}$. In general, a higher PSNR indicates better perceptive quality of the synthesized CT image. Note that in order to compare quantitative results with other methods (Huynh et al., 2016) in terms of MAE, we rescale the normalized intensity value back by reversing normalization step.

### 3.3 Role of Embedding Block

It is necessary to verify the importance of the embedding blocks in DECNN. The *embedding blocks* can be flexibly appended to the *transform stage*. With 5 simple convolutional layers *prior to* the first embedding block, we show the PSNR and MAE changes *with respect to* different numbers of embedding blocks in Fig. 3 (a-d), when applied to the brain and prostate datasets, respectively. Specifically, "Ebd $k$" indicates $k$ embedding blocks used in DECNN, where the overall number of the convolutional layers for transforming feature maps is $(3k + 5)$ as in Fig. 2(c).

From Fig. 3 (a) and (c), we can observe that the PSNR measures increase quickly to the best performances with more embedding blocks and thus deeper networks. However, the performance starts to drop once the DECNN model becomes too deep to be well trained, partially due to the small size of each dataset. Specifically, the highest average PSNR metric occurs at "Ebd 2" on the brain dataset, while at "Ebd 4" on the prostate dataset, respectively. The use of less *embedding blocks* for the brain dataset is mainly because the brain CT images are relatively easy to synthesize. On the other hand, a deeper network (with more *embedding blocks*) is needed to map from MR to CT, due to more pelvic anatomical details in the prostate CT images. We also show the comparison with MAE measurement in Fig. 3 (b) and (d). Similar observations can be seen that "Ebd 2" and "Ebd 4" also get the lowest value on brain dataset and prostate dataset respectively, which shows they are the optimal setting.

The DECNN model is further compared to the CNN model without embedding of the feature maps. Since we apply our proposed method with "Ebd 2" to the brain dataset, we compare it to the CNN model with the same number of convolutional layers (i.e., 11 layers in the transform stage), which is referred to as "11-layer CNN" in Fig. 3. The average PSNR is 27.3dB with embedding, while 26.7dB without embedding. Similarly, for the prostate dataset, we compare our proposed method with "Ebd 4" to "17-layer CNN". The average PSNR of these two models is 33.5dB and 33.1dB, respectively. The improvements of our proposed DECNN are statistically significant for both datasets (with $p$-values of 0.0022 and 0.0086 in paired $t$-tests, respectively). The above results show that, compared with CNN with the same depth, our DECNN can encode the mapping from MR to CT more effectively.

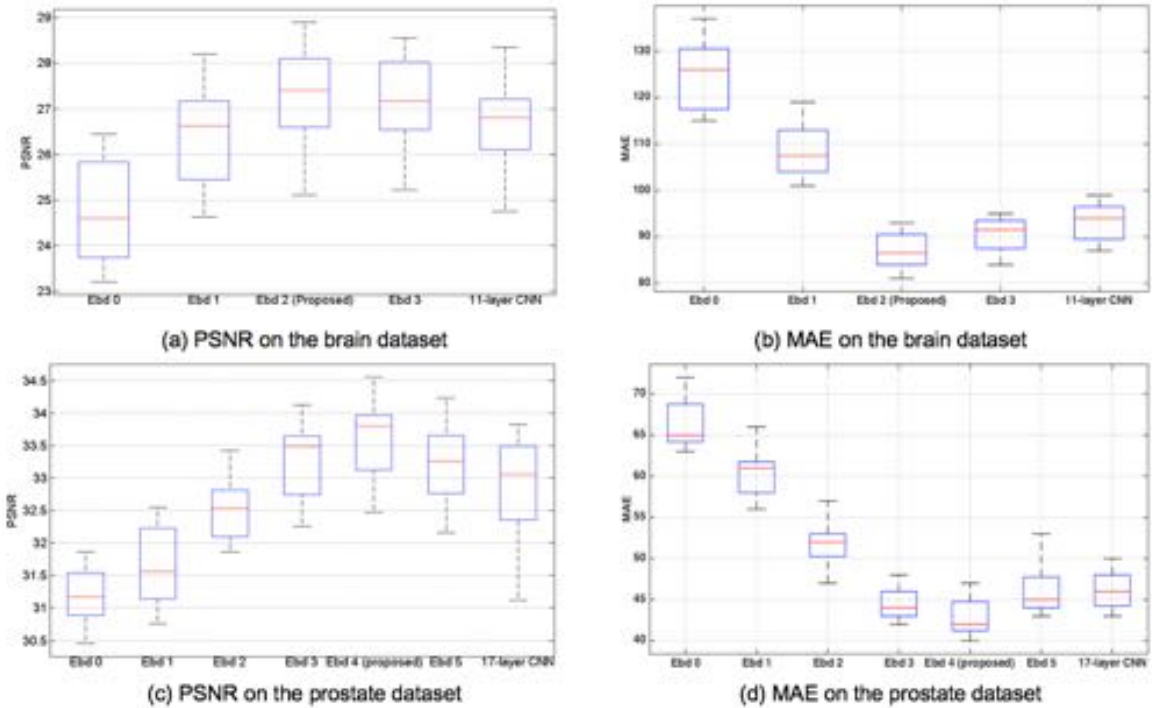

**Fig. 3.** Comparisons of the performances by different networks, measured with PSNR and MAE, for (a-b) brain dataset and (c-d) prostate dataset. Here, "Ebd $k$" indicates $k$ embedding blocks used in DECNN, where the number of the convolutional layers for transforming feature maps is $(3k + 5)$. The comparison methods are referred to as the CNN models without embedding, but with the same respective numbers of convolutional layers, such as "11-layer CNN" for the brain dataset and "17-layer CNN" for the prostate dataset.

The use of embedding blocks is also helpful for easier convergence of the network, which is fundamentally important when training the DECNN model for practical usage. In Fig. 4 (a) and (b), we show the testing performance changes in PSNR, when training the network with more epochs of patches. On both brain and prostate datasets, DECNN converges faster than the CNN model without embedding, even using the same number of the convolutional layers. In particular, while the depth becomes deeper to handle for the prostate images than the brain images, DECNN leads to a wider margin ahead of the CNN model without embedding. All the observations above imply that the use of embedding strategy upon the feature maps is effective in the optimization of parameters in the convolutional network, especially when the deep network is used to encode the highly complex end-to-end mapping.

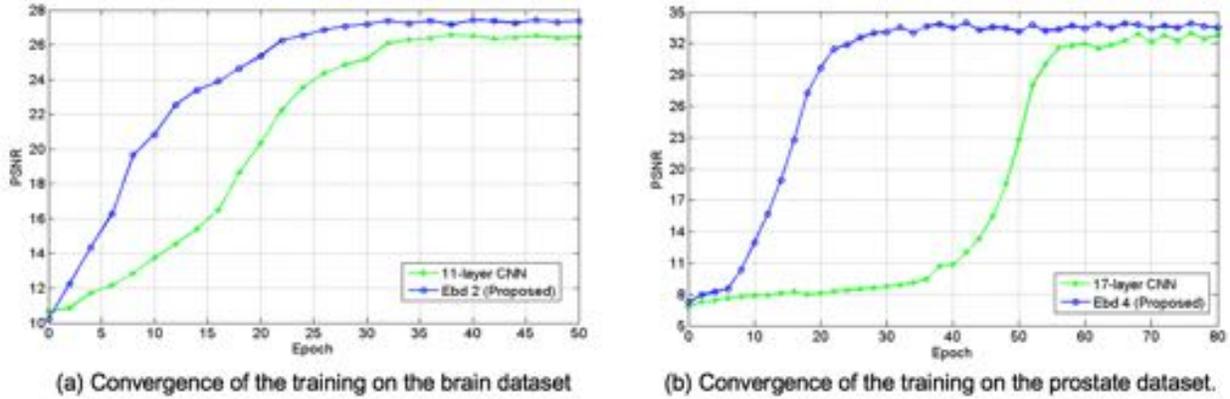

**Fig. 4.** The changes of the testing performance *with respect to* the increasing epochs of the training samples are plotted in (a) and (b) for brain and prostate datasets, respectively, to demonstrate the convergence of the training of DECNN. The comparison methods are referred to as the CNN models without embedding, but with the same respective numbers of convolutional layers, such as "11-layer CNN' for the brain dataset and "17-layer CNN" for the prostate dataset.

The CT estimation results by using different 'Ebd *i*' are shown in Fig. 5 for brain dataset, and Fig. 6 for prostate dataset. In Fig 5, the first and the second columns show the MR image input to the DECNN and the ground-truth CT image, respectively, in the axial, sagittal and coronal views. 'Ebd 0' can estimate the contour of the bone, but also generate many artifacts in the synthesized CT image. With the use of one embedding block, the result of 'Ebd 1' becomes better. When using 'Ebd 2', we can get the most satisfying result compared to the ground-truth. Similar observations can also be found for the prostate dataset in Fig. 6. Note that more embedding blocks are needed for synthesizing the prostate CT images, as the MR-to-CT mapping is more complex for the pelvic region.

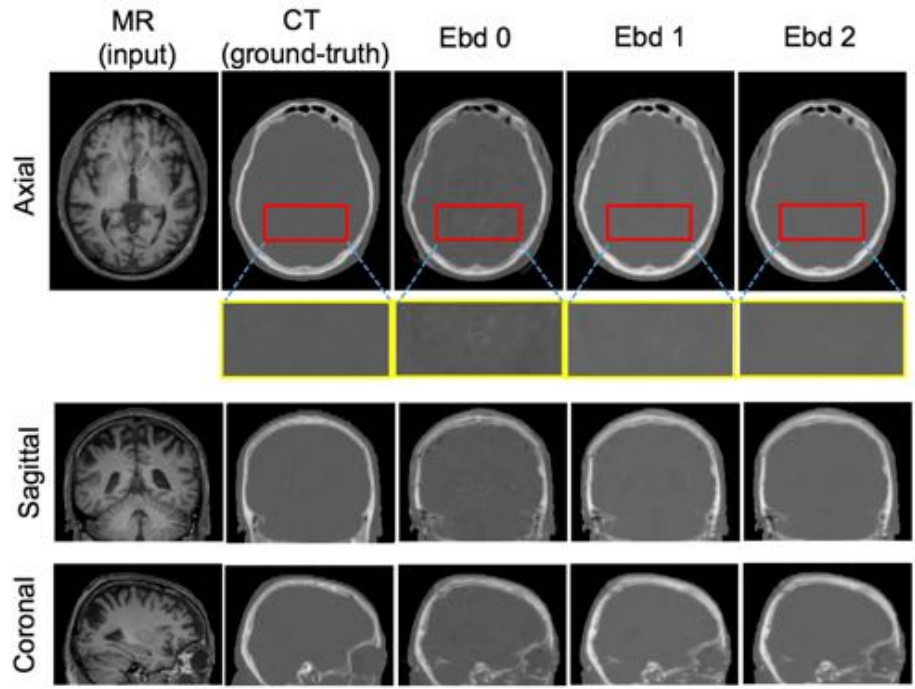

**Fig. 5.** The brain CT images synthesized using our DECNN model with different numbers of embedding blocks, i.e., 0 (middle), 1 (last second column), and 2 (last column).

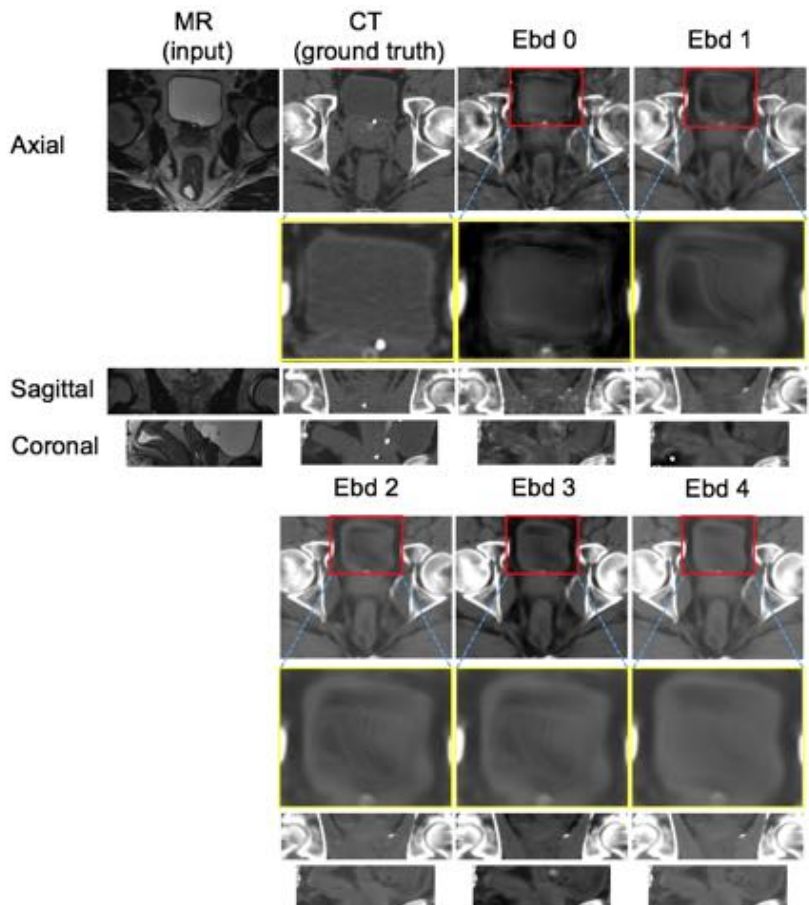

**Fig. 6.** The prostate CT images synthesized using our DECNN model with different numbers of embedding blocks, i.e., 0 and 1 in the last two columns of the upper panel, and 2 to 4 in the lower panel.

### 3.4 Quasi-3D Mapping

In DECNN, we implement the input as 3 consecutive axial slices from MR. The 3 MR slices are jointly considered as part of the feature maps since the first convolutional layer in the network. The output consists of 3 axial slices that correspond to the input. We synthesize every 3 consecutive axial slices for the subject CT image, and combine the outputs into the final result by simple averaging. We regard this joint synthesis of the 3 consecutive slices as *quasi-3D mapping*. Examples of the synthesized brain CT images are denoted by "×3 slices" for the implemented DECNN, and compared with the cases of fewer ("×1 slice") or more ("×5 slices") input slice(s) in Fig. 7.

The comparisons are two-fold. *First*, concerning the quality within the axial plane (c.f. the red boxes in the first row of the figure), "×3 slices" is comparable to "×5 slices" and better than "×1 slice". *Second*, the outputs of "×3 slices" and "×5 slices" are relatively continuous in the superior-inferior direction (c.f. both the yellow and the purple boxes highlighted on the sagittal and the coronal views). This implies that, with more than one axial slices as input, DECNN models the MR-to-CT mapping in the *quasi-3D* way. The mapping benefits *not only* the intra-slice quality, *but also* the inter-slice continuity. Fewer artifacts are observable within and between the axial planes. We note that, with "×5 slices", the outputs are mostly comparable to "×3 slices" visually (as shown in Fig. 7) and quantitatively (in terms of average PSNR: 27.4 dB for "×5 slices" and 27.3 dB for "×3 slices" on the brain dataset). Accordingly, we adopt the "×3 slices" setting for the proposed DECNN, to reduce the complexity of the computation.

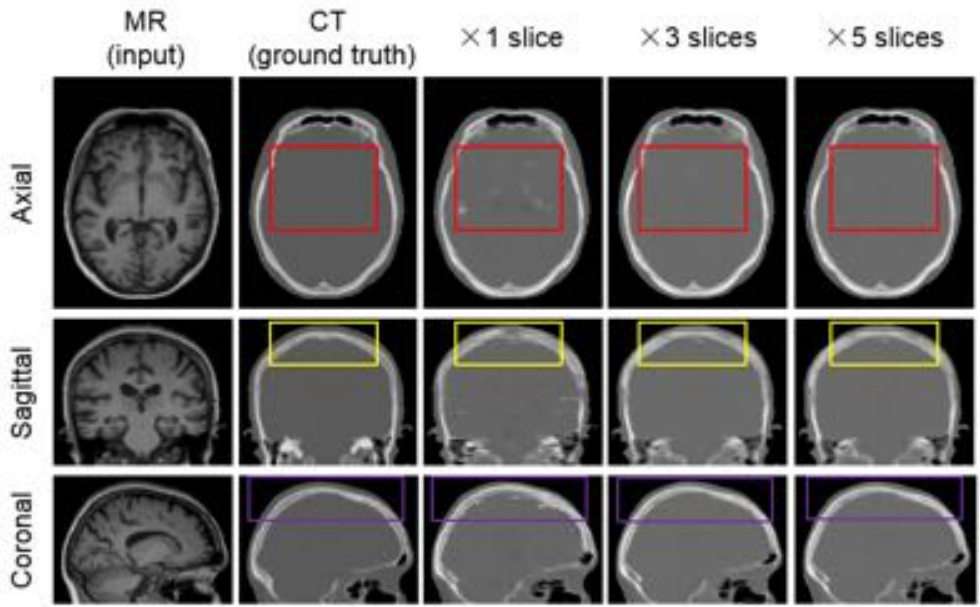

**Fig. 7.** The synthesized CT brain images using 1, 3 and 5 axial slices as inputs to DECNN, which are designated as "×1 slice", "×3 slices" and "×5 slices", respectively. The results in the axial, sagittal and coronal views are presented.

### 3.5 Comparisons with State-of-the-Art Methods

In this section, we conduct comprehensive comparisons between our DECNN and state-of-the-art methods, including the atlas-based method ("Atlas"), the sparse representation method ("SR"), and the random forest with the auto-context model ("RF+ACM") (Huynh et al., 2016). The results of the CNN models without embedding, which are designated by "CNN" for short, are also summarized. A 3D CNN result termed as "FCN" is also provided for comparison (Nie et al., 2016). While there are many variants of atlas-based (Burgos et al., 2014; Hofmann et al., 2008; Kops and Herzog, 2007) and sparse representation based methods (Iglesias et al., 2013; Roy et al., 2011; Roy et al., 2010; Wright et al., 2009), we only chose one standard implementation for these two approaches for demonstration purpose:

- "Atlas": When testing a subject, all other training images are used as atlases; that is, we are using leave-one-out cross validation here. To synthesize the CT image of a test subject, the MR image of each training atlas is first aligned to the subject MR image. The deformation field will be used to warp the CT image of each training atlas. Then, we can have the final CT synthesis by fusing the warped CT images of all training atlases.
- "SR": When all the training atlases are warped to the subject MR image, a non-local sparse representation is carried out. That is, a small MR patch in the subject MR image is represented as a sparse liner combination of non-local MR patches from the aligned training atlas MR images. Then, the representation coefficients are applied to the respective aligned training atlas CT images to obtain the final synthesized CT.

- "RF+ACM": This is a learning-based method which uses the structured random forest to directly estimate a CT image as a structured output from the input MR image. Each leaf node of the trees in the random forest yields an estimate, which can be integrated for the final output. Moreover, the auto-context model is used for further refining the synthesized CT image.

The detailed MAE and PSNR measures are shown in Table 1 and Table 2, which correspond to the brain and prostate datasets, respectively. The results of the existing methods can be obtained from (Huynh et al., 2016). DECNN and CNN clearly outperform the other three methods under comparison, which imply the powerfulness of deep convolutional networks in mapping from MR to CT. Further, with the same depth (or the number of convolutional layers), the proposed DECNN is superior to the CNN model without embedding, by producing higher PSNRs and lower MAEs. When compared to FCN model, our DECNN results in better predictions on brain dataset and equal performance on prostate dataset. In general, our DECNN is able to attain the highest performance among state-of-the-art methods.

Visual comparisons for different methods are also provided in Fig. 8 for brain dataset and Fig. 9 for prostate dataset. For each dataset, we give two exemplar subjects. We can observe that the CT images synthesized by DECNN are more similar to the ground truth in both datasets. The structural similarity is high, while more anatomical details are preserved in the results of DECNN. The CT images synthesized by "Atlas" and "RF+ACM", however, are relatively fuzzy (c.f. the bone boundaries highlighted in the figure). The CT images synthesized by CNN has much noise in it (see Fig. 8). The statistic results on prostate dataset are almost the same in Table 2, 33.5 (ours) vs 33.4 (Nie et al.). But the visualized result by FCN is not as good as ours, the zoomed in part in Fig. 8 shows that the bone is not accurately synthesized, also the zoomed part in Fig. 9 shows that the bone boundary is not clear. Another advantage for our DECNN compared to FCN is that we don't need any post-processing in testing stage, FCN needs to padding the input image to get the desired output size due to their architecture that makes their output size smaller than the input size. Our DECNN is more efficient compared to FCN method as our output size is equal to input size. Therefore, the proposed DECNN method yields better perceptive quality of the synthesized CT images than other state-of-the-art methods.

DECNN is featured by the end-to-end mapping capability that allows us to process each test subject very efficiently. On the contrary, the conventional methods are usually time-consuming. For example, "RF+ACM" synthesizes the CT images by extracting features first and then processing through random forest regressors with auto-context. Even though the PSNRs of "RF+ACM" are relatively higher than both "Atlas" and "SR" method, its computation efficiency still falls short significantly compared to "CNN" and "DECNN". Specifically, "RF+ACM" takes about 20 minutes to synthesize a CT image for the test subject. With "DECNN", we can complete the computation of a brain subject (of the size 234×181×149) in 46 seconds, and a prostate subject (of the size 153×193×50) in 28 seconds. The high speed performance provides the potentials for clinic applications of our method. The above experiments are carried out on a computer with an Intel Core i7 4.00GHz processor, 16 GB RAM, and an NVIDIA GeForce GTX Titan X GPU.

Table 1. The MAE (HU) and PSNR (dB) measures of the CT synthesis results by different methods on the *brain* dataset.

| Brain | Atlas | | SR | | RF+ACM | | CNN | | FCN | | DECNN | |
|---|---|---|---|---|---|---|---|---|---|---|---|---|
| | MAE | PSNR | MAE | PSNR | MAE | PSNR | MAE | PSNR | MAE | PSNR | MAE | PSNR |
| Mean | 169.5 | 20.9 | 166.3 | 21.1 | 99.9 | 26.3 | 93.6 | 26.7 | 88.9 | 27.1 | **85.4** | **27.3** |
| Std. | ±35.7 | ±1.6 | ±37.6 | ±1.7 | ±14.2 | ±1.4 | ±11.2 | ±1.0 | ±10.6 | ±1.3 | ±9.24 | ±1.1 |
| Median | 169.9 | 20.5 | 164.3 | 21.1 | 97.6 | 26.3 | 94.1 | 26.8 | 89.6 | 26.9 | **86.1** | **27.6** |

Table 2. The MAE (HU) and PSNR (dB) measures of the CT synthesis results by different methods on the *prostate* dataset.

| Prostate | Atlas | | SR | | RF+ACM | | CNN | | FCN | | DECNN | |
|---|---|---|---|---|---|---|---|---|---|---|---|---|
| | MAE | PSNR | MAE | PSNR | MAE | PSNR | MAE | PSNR | MAE | PSNR | MAE | PSNR |
| Mean | 64.6 | 29.1 | 54.3 | 30.4 | 48.1 | 32.1 | 45.3 | 33.1 | **42.4** | 33.4 | 42.5 | **33.5** |
| Std. | ±6.6 | ±2.0 | ±10.0 | ±2.6 | ±4.6 | ±0.9 | ±4.2 | ±0.9 | ±5.1 | ±1.1 | ±3.1 | ±0.8 |
| Median | 65.9 | 29.8 | 54.0 | 31.3 | 48.3 | 31.8 | 44.9 | 33.0 | 42.6 | 33.2 | **42.3** | **33.8** |

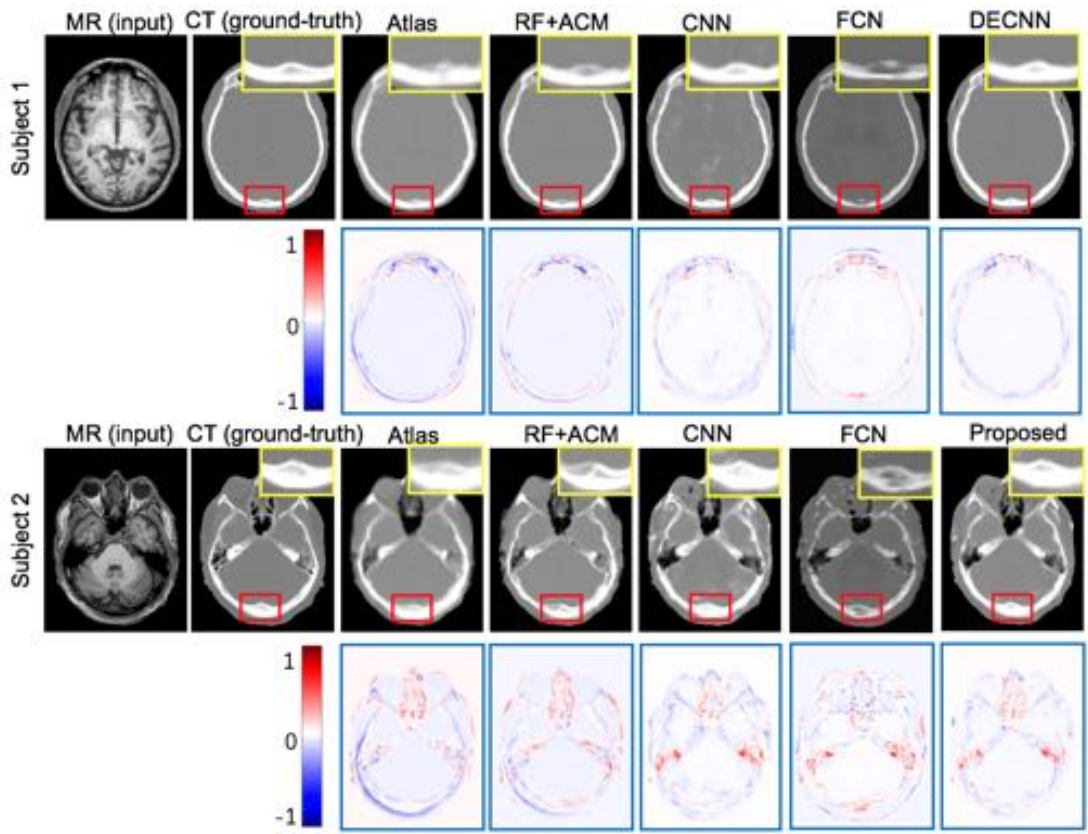

**Fig. 8.** Comparisons with state-of-the-art methods on two exemplar *brain* images. The first row in each panel shows the synthesized CT images by different methods. The second rows show the residuals between the synthesis and the ground truth.

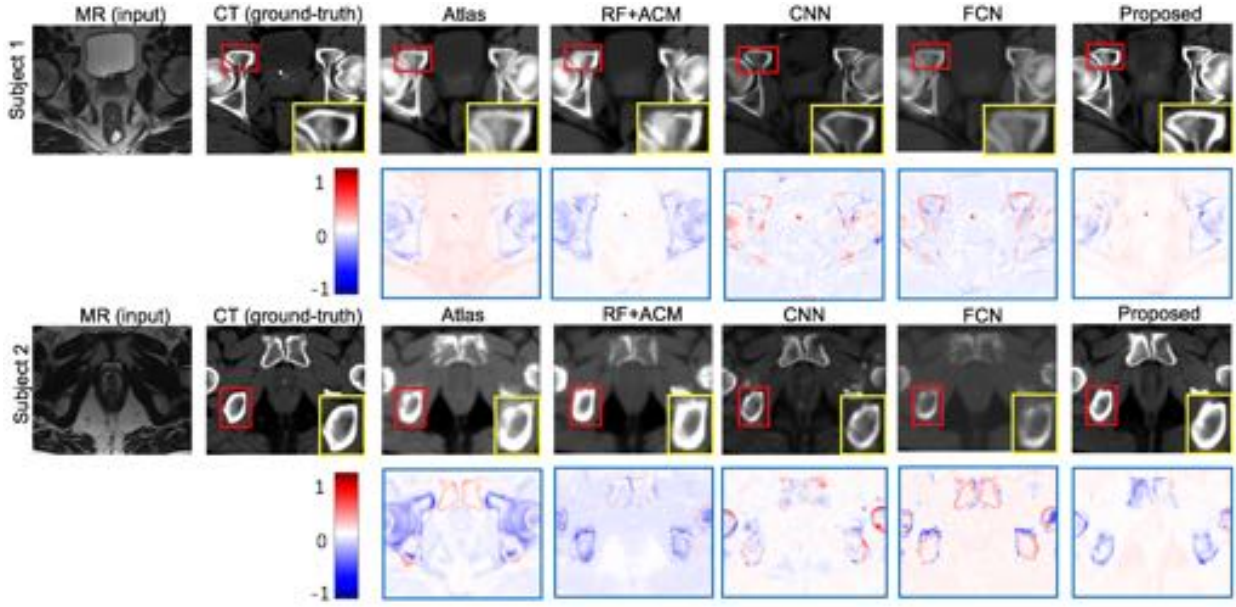

**Fig. 9.** Comparisons with state-of-the-art methods on two exemplar *prostate* images. The first row in each panel shows the synthesized CT images by different methods. The second rows show the residuals between the synthesis and the ground truth.

### 3.6 Evolution of Feature Maps

For further understanding the mapping from MR image to the corresponding CT image, we analyze the evolving feature maps after several convolutional layers in DECNN. First, we use the structural mutual information (SMI) metric to compute the similarity between the feature map $A$ and the ground-truth CT image $B$ as we want to learn feature maps related to final predicted CT image:

$$SMI = H(A.* G(A)) + H(B.* G(B)) - H(A.* G(A), B.* G(B)) \qquad (7)$$

where $G(A)$ and $G(B)$ are the gradient maps of $A$ and $B$, respectively. $H(\cdot)$ computes the joint entropy of the input images. We sort these feature maps in descending order by the SMI values, then select 8 feature maps (column) for brain dataset and 7 feature maps (column) for prostate dataset. The exemplar feature maps are selected for visualization in Fig. 10 for the brain dataset, and Fig. 11 for the prostate dataset. For the brain dataset, in addition to the 3 consecutive MR slices as input and the 3 CT slices as ground-truth output in Fig. 10 (a), we also show the feature maps produced by individual layers of the adopted 'Ebd 2' model. In particular, the layers of 'Conv 1', 'Conv 3', 'Conv 5', 'Conv 8' and 'Conv 11' are selected for visualization, where 'Conv 1-5' indicate the convolutional layers prior to the first embedding block, and 'Conv 8' and 'Conv 11' are in the end of the two embedding blocks, respectively.

In 'Conv 1', the original MR slices are transformed to a series of feature maps, each of which aims to synthesize brain CT image. That is, the soft tissue structures are discarded while the bone regions of the brain boundaries are focused. Note, the original MR slices are decomposed into many feature maps according to intensity values, and the feature maps with very dark bone regions will be preserved as they are associated with the bone regions of the brain CT image (the last few columns in 'Conv1' of Fig. 10 (b)). Their shape and location are corresponding to that of ground truth CT image, but the intensity values in the bone mask are not changed compared to the input MR (both dark). So the latter layers are mainly to seek the intensity mapping on this corresponding region. Specially, in 'Conv 3', the feature map (indicated in the green box) has becomes white in the bone mask region. This verifies the transform mapping from MR space to CT space are almost learned. In the end of transform block (Conv 5), we can see more CT-like feature maps (as indicated by the blue box). After applying the embedding block, the CT-related feature maps are getting much meaningful and clear (as indicated by both yellow box and red box). Specifically, the feature map in the red box is more satisfying, since there is less noise in the middle of the feature map and all the information in this feature map is more related to the CT image.

Similar observations can be also found on the prostate dataset in Fig. 11 (a-b). But, there are a few differences compared to the brain images, i.e., the decomposition way in earlier convolutional layers, as well as using more embedding blocks for prostate such as using the 'Ebd 4 (proposed)' model. For the case of brain image, the network decomposes MR image according to its intensity value while ignoring structure information in the middle of the brain. However, for the case of prostate CT image, there is also complex boundary information between different soft tissues in the CT image. Therefore, all the decomposed feature maps preserve boundary information from the MR image (as shown in the first row 'Conv 1' in Fig. 10 (a)). Also, prostate CT image has more *structural details* compared to brain CT image. Therefore, it needs more embedding blocks to complete the MR-to-CT mapping. Similarly, feature maps in the latter convolutional layers are getting more related to CT image, and details are becoming clearer compared to earlier layers. For example, both feature maps in the blue box of row 'Conv 14' and in the red box of row 'Conv 17' are related to the bone structures, while the boundary in 'Conv 17' is much clearer than the one in 'Conv 14', which can yield more accurate CT image prediction results.

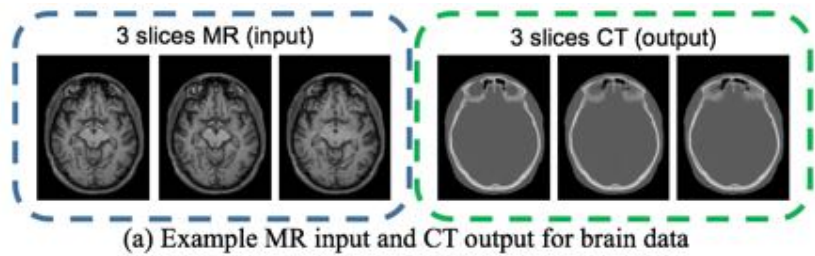

(a) Example MR input and CT output for brain data

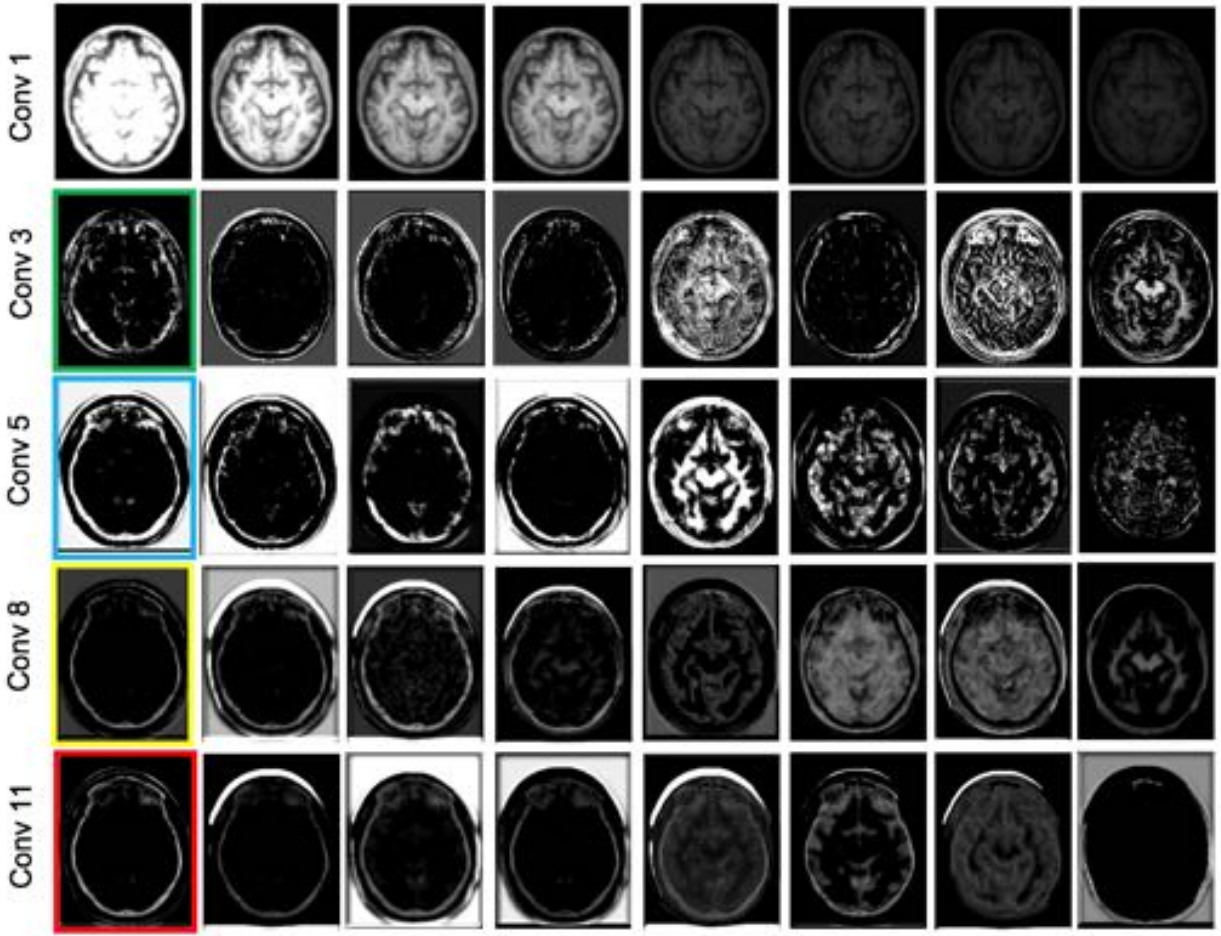

(b) Example feature maps of different layers for brain data

**Fig. 10.** Example feature maps in each convolutional layer for the case of synthesizing brain CT image. These feature maps are sort by descending order according to their SMI values. The most CT-like feature map in the convolution layer is chosen in the colored boxes.

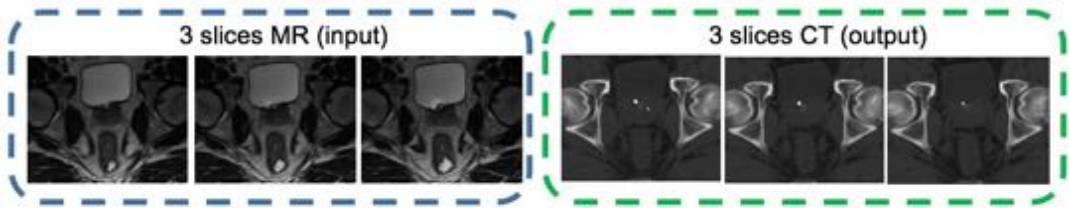

(a) Example MR input and CT output for prostate data

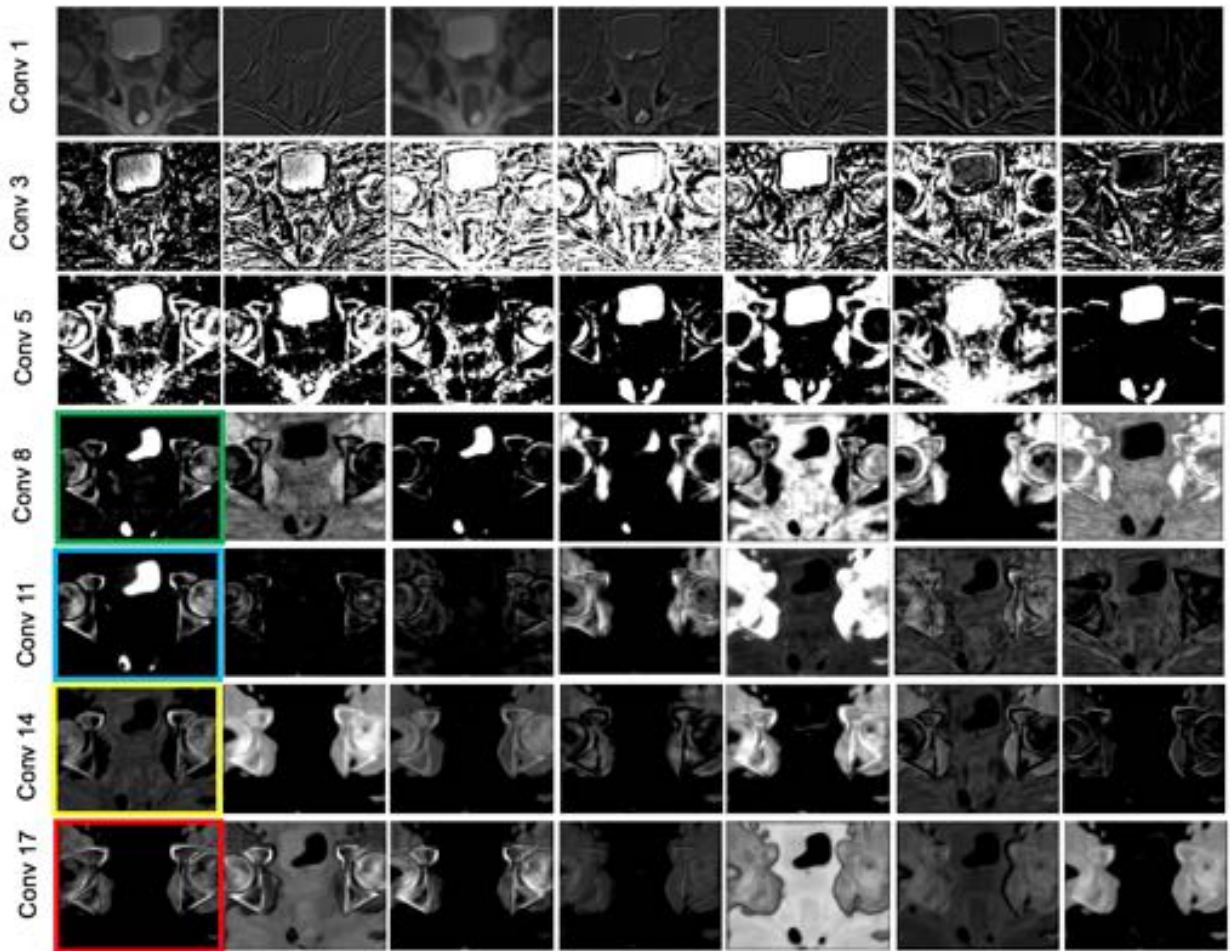

(b) Example feature maps of different layers for prostate data

**Fig. 11.** Examples feature maps in each convolutional layer for the case of synthesizing prostate CT image. These feature maps are sort by descending order according to their SMI values. The most CT-like feature map in the convolution layer is chosen in the colored boxes.

All obtained examples show how the information in MR image can be gradually transformed to CT image through serial layers. Also, these examples show why prostate dataset needs more embedding blocks for CT synthesis. This is mainly because more structural information on the boundaries of soft tissues need to be estimated in the prostate CT than in the brain CT.

In order to demonstrate the difference of feature maps learned by our proposed DECNN and basic CNN, we show some feature maps from the last layer of these two models in Fig. 12. (a) and (b) show feature maps learned by CNN and DECNN on brain dataset, respectively. We find that in the feature maps in (a), there are unrelated patterns (the white part in the center) that are not related to CT image. We argue that these are redundant information for the desired CT image. When comes to feature maps (Fig. 12 (b)) from DECNN, they are mostly related to CT image and they have more clear boundary. Similar observations could be found on prostate dataset (Fig. 12 (c) and (d)).

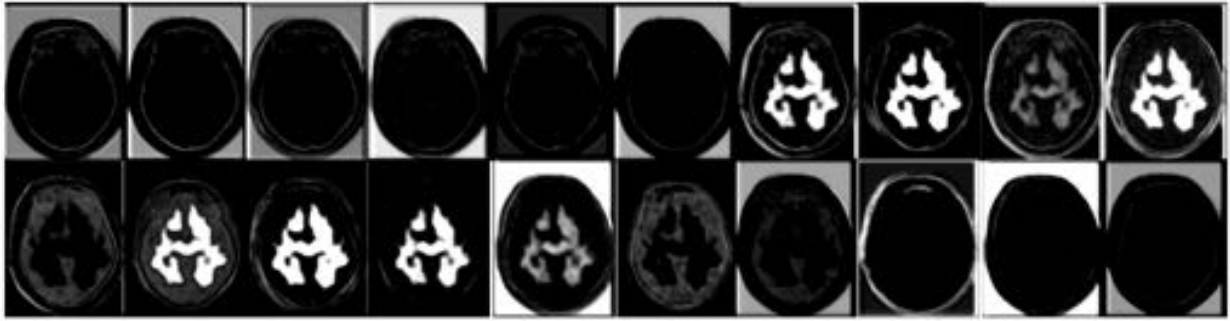
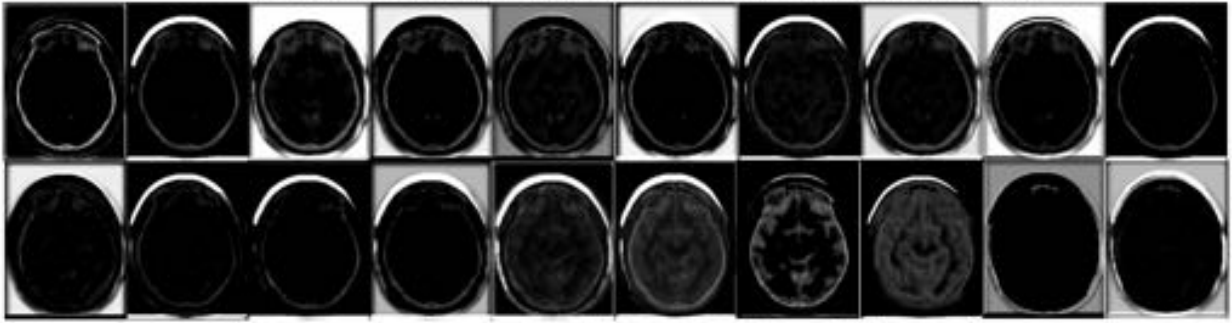
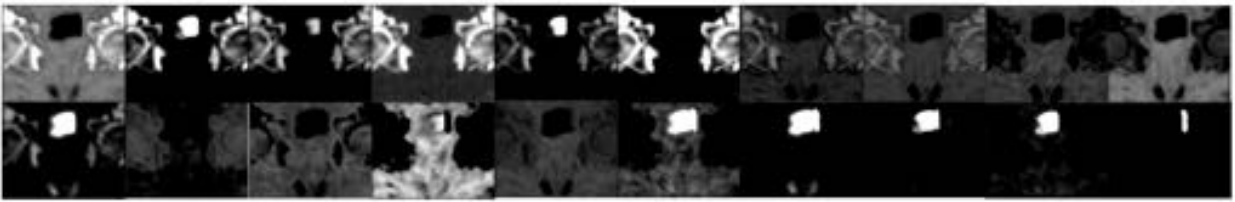
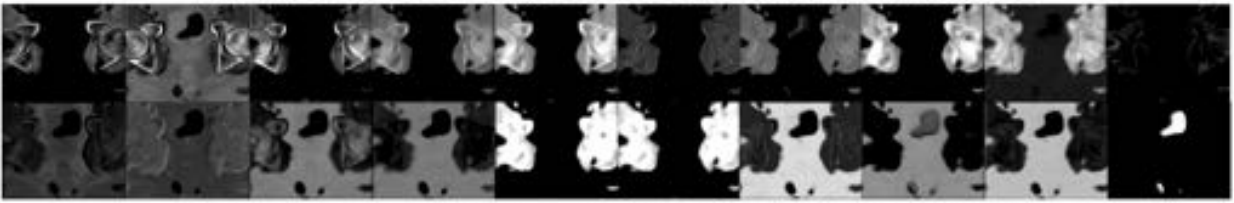

**Fig. 12.** Feature maps learned from last layer by CNN and DECNN on both brain and prostate dataset.

## 4 Discussion

We have presented a novel MR-to-CT mapping method for different modality transform on both brain and prostate dataset. Compared with the traditional learning based method, our DECNN model not only achieve best synthesis result but also perform several times or even orders of magnitude faster than them in testing stage. There are also some limitations for our method. Firstly, our method costs lots of time for training, which takes 2-3 days to get a model, while traditional methods mostly spend less than a day. Future work on this aspect should consider the training efficiency on how to reduce network capacity without sacrificing the accuracy. Secondly, the noise in the training dataset could affect the performance in testing stage. As shown in Fig. 5 coronal view, there are some bone intensity noise in the predicted result. Also the slight misalignment for preparing paired MR-CT images for training DECNN could also cause inaccurate CT prediction. In the future we may also focus on how to train a robust model with such dataset to handle with this distortion.

In current work we perform MR-to-CT mapping in bone window on brain dataset as we just focus on bone part. In the future, we would extend the work on all range of intensity values, including the soft tissue contrast. Regarding to the prostate dataset, we mainly focus on three organs: prostate, bladder and rectum. The remaining parts are not our concerns currently. When we need to synthesize a new whole image for clinically relevant applications (attenuation correction or radiotherapy treatment planning), we could perform a linear registration for the testing image to align to our training image, then crop the interested region including these three organs and do the CT synthesis. This should satisfy the demands for clinically applications as well. But this needs extra post-processing. We would also explore the robustness of this method on the whole pelvic image for convenience of clinically applications.

## 5 Conclusion

In this paper, we propose a novel DECNN model to synthesize the CT image from the T1-weighted MR image. Deep learning is well known for its capability in encoding the highly complex mapping between two different image spaces. The embedding block, which embeds the tentative CT estimation into the flow of the feature maps in deep learning, is integrated with CNN in our work. Thus, we derive DECNN that can transform the embedded feature maps forward and reconstruct better CT synthesis results in the end. Our experiments also prove that our proposed embedding block can effectively contribute to more accurate modeling of the MR-to-CT mapping, as well as faster convergence of the training of the deep network. Meanwhile, the overall performance of DECNN, measured by both the quality of the synthesized CT images and the running time, is superior to the state-of-the-art methods.

Note, our work provides a generalized deep learning solution that is not restricted to the MR-to-CT synthesis only. Moreover, we adopt the end-to-end mapping, such that the resulted CT image can be exactly the same size as the input MR image. In this way, the architecture of our DECNN ensures high efficiency to process the test images. In the future, we will further investigate the performance of DECNN when applied to other image synthesis tasks, thus promoting more usage in the real clinical applications (e.g. dose planning in the prostate radiation and attenuation correction for PET imaging in PET/CT scanner).